\documentclass[sigconf]{acmart}
\renewcommand\footnotetextcopyrightpermission[1]{} 
\AtBeginDocument{%
  \providecommand\BibTeX{{%
    \normalfont B\kern-0.5em{\scshape i\kern-0.25em b}\kern-0.8em\TeX}}}

\setcopyright{acmcopyright}
\copyrightyear{2024}
\acmYear{2024}

\acmConference[Preprint]{Preprint}{2024}{Sydney}
\usepackage{rotating}
\usepackage{multirow}
\usepackage{xcolor}
\usepackage{bbm}

\begin{document}

\title{Prompt Mining for Language-based Human Mobility Forecasting}

\author{Hao Xue}
\authornote{Both authors contributed equally to this research.}
\email{hao.xue1@unsw.edu.au}
\affiliation{%
  \institution{University of New South Wales}
  \city{Sydney}
  \state{NSW}
  \country{Australia}
}

\author{Tianye Tang}
\email{tianye.tang@student.unsw.edu.au}
\authornotemark[1]
\affiliation{%
  \institution{University of New South Wales}
  \city{Sydney}
  \state{NSW}
  \country{Australia}
}
\author{Ali Payani}
\email{apayani@cisco.com}
\affiliation{%
  \institution{Cisco Systems Inc.}
  \country{USA}
}
\author{Flora D. Salim}
\email{flora.salim@unsw.edu.au}
\affiliation{%
  \institution{University of New South Wales}
  \city{Sydney}
  \state{NSW}
  \country{Australia}
}

\renewcommand{\shortauthors}{Xue et al.}
\newcommand{\ie}{\textit{i.e.}}
\newcommand{\eg}{\textit{e.g.}}
\begin{abstract}
  With the advancement of large language models, language-based forecasting has recently emerged as an innovative approach for predicting human mobility patterns. The core idea is to use prompts to transform the raw mobility data given as numerical values into natural language sentences so that the language models can be leveraged to generate the description for future observations. However, previous studies have only employed fixed and manually designed templates to transform numerical values into sentences. Since the forecasting performance of language models heavily relies on prompts, using fixed templates for prompting may limit the forecasting capability of language models. In this paper, we propose a novel framework for prompt mining in language-based mobility forecasting, aiming to explore diverse prompt design strategies. Specifically, the framework includes a prompt generation stage based on the information entropy of prompts and a prompt refinement stage to integrate mechanisms such as the chain of thought. Experimental results on real-world large-scale data demonstrate the superiority of generated prompts from our prompt mining pipeline. Additionally, the comparison of different prompt variants shows that the proposed prompt refinement process is effective. Our study presents a promising direction for further advancing language-based mobility forecasting.
\end{abstract}



\maketitle

\section{Introduction}
The forecasting of human mobility plays a crucial role in various domains, including urban planning, transportation management, public health preparedness, and disaster response. Accurate predictions of movement patterns and trends can facilitate better location-based services, proactive decision-making, optimize resource allocation, and enhance overall societal resilience.
Traditional statistical forecasting methods~\cite{calabrese2010human,qiao2018hybrid} often rely on numerical models, but they might fall short in capturing the complex human behaviour in real-world scenarios. In an effort to capture these complex patterns, more advanced deep learning forecasting methods~\cite{feng2018deepmove,sun2020go,yin2023mtmgnn}, particularly Transformer-based approaches\cite{xue2021termcast,hong2022you,xue2021mobtcast}, have emerged. However, these methods often require very complicated model architectures.

\begin{figure*}
    \centering
    \includegraphics[width=0.95\textwidth]{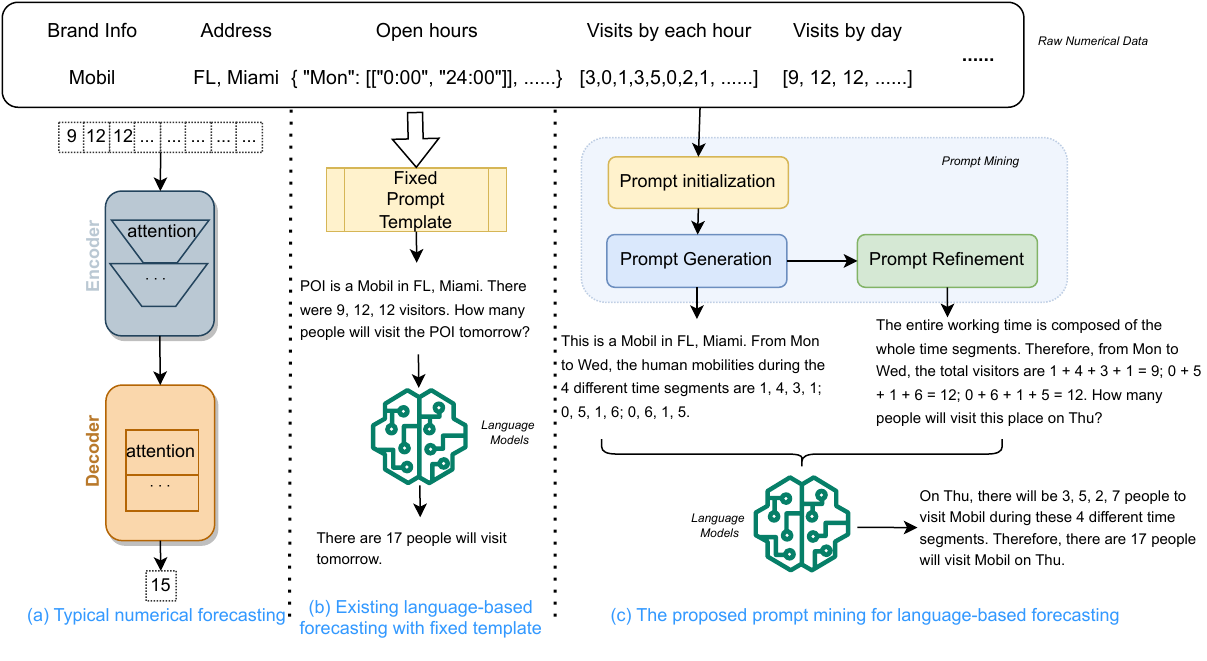}
    \caption{The conceptual comparison of: (a) numerical forecasting, (b) language-based forecasting with a fixed template, (c) our proposed prompt mining process.}
    \label{fig:intro}
\end{figure*}

Recently, language-based forecasting approaches~\cite{xue2022translating,xue2022leveraging,xue2023promptcast} have demonstrated another line of work and emerged as a promising alternative, harnessing the power of natural language processing (NLP), language generation frameworks, and advanced pre-trained language models to generate forecasts. These methods present a new paradigm (as shown in Figure~\ref{fig:intro} (b)) for forecasting human mobility.
By transforming numerical data into natural language sentences, these methods make it possible to use existing available language models to comprehend and predict human mobility patterns.
This shift towards language-based forecasting has opened up new avenues for enhancing predictive capabilities. Language models, such as Bert~\cite{bert} and its successors, have demonstrated remarkable proficiency in understanding and generating human language. Leveraging their capabilities for mobility forecasting offers the potential to achieve accurate predictions while also reducing the need of designing complex specific forecasting models.

However, a key challenge in language-based mobility forecasting lies in the design of effective prompts, that is, how to transform the numerical mobility data and associated auxiliary information into sentences for language models.
These prompts serve as the bridge between raw numerical data and natural language descriptions, shaping how well language models can capture the underlying mobility patterns.
Generally, different prompts can lead to varying language generation performance, consequently impacting the accuracy of the forecasting.
Taking \textit{ChatGPT} as an example, even for the same topic, using different prompts as inputs will largely result in different responses.
Hence, although using fixed manually designed templates for prompt generation as in existing work~\cite{xue2022translating,xue2022leveraging,xue2023promptcast} is simple and straightforward, inadequate prompt template exploration can lead to inaccurate forecasts

To address these limitations, we propose a novel prompt mining framework for the language-based mobility forecasting task (Figure~\ref{fig:intro} (c)). Our framework is a multi-stage pipeline with prompt initialization, prompt generation, and prompt refinement stages.
Specifically, in the prompt generation stage, we introduce a Prompt Quality Evaluator that evaluates the quality of generated prompts based on a combination of heuristic classifier rules and prompt entropy. By doing so, we can quantitatively evaluate the quality of prompts, which enables the prompt generation model to receive feedback for generating high-quality prompts during training.
The prompt refinement stage aims to enhance the generated prompts through a series of sophisticated mechanisms including noise reduction, integration of a chain of thought, and the application of information gain-based temporal segmentation on mobility data. These strategies collectively contribute to generating more refined and contextually relevant prompts.
Through extensive experiments with real-world large-scale mobility data, we show the effectiveness of our prompt mining approach. Our results showcase the superiority of generated prompts compared to traditional fixed templates. 
In summary, our work has 3 main contributions: 
\begin{itemize}
    \item We propose a novel prompt mining framework that addresses the limitations of relying on fixed templates in existing methods. The chain of thought consideration is also integrated into the prompt mining pipeline.
    \item In our framework, the concept of information entropy is explored in both the semantics of the prompt in the prompt generation stage as well as the distribution of the mobility data in the prompt refinement stage.
    \item We conduct extensive experiments on real-world mobility data to empirically validate the efficacy of our proposed method. Our mined prompt variants show good performance under the language-based mobility forecasting setting. 
\end{itemize}

\section{Related Work}

\subsection{Numerical Time Series Forecasting}
Human mobility forecasting is often conceptualized as a special case within the broader scope of general time series forecasting. Consequently, methodologies developed for general time series forecasting are frequently applied to this domain.
Modern time series forecasting techniques heavily rely on deep learning, primarily utilizing Recurrent Neural Network (RNN) architectures such as Long Short Term Memory (LSTM) networks~\cite{hochreiter1997long} and Gated Recurrent Units (GRU)~\cite{chung2014empirical}.
Examples within this RNN-based framework for predicting sequential human behavior include ST-RNN~\cite{liu2016predicting} and DeepMove~\cite{feng2018deepmove}.
Motivated by the recent success of applying Transformer architecture~\cite{vaswani2017attention} in modeling nature language sequences, it has also been applied for human mobility forecasting~\cite{xue2021termcast,xue2021mobtcast} and general time series forecasting tasks~\cite{zhou2021informer,xu2021autoformer,FEDformer}.

These methodologies, whether based on RNNs or advanced Transformer architectures, typically adopt a sequence-to-sequence approach, where historical numerical observations form the input sequence to predict future outcomes. 
As the need to incorporate semantic information (\eg, a POI is a restaurant or a shop) arises to further improve forecasting performance, the complexity of model architectures tends to increase.

\subsection{Language-based Forecasting}
Efforts to integrate semantic information more directly and effectively have led to the introduction of shaping mobility forecasting as a language generation problem.
Xue et al.~\cite{xue2022translating} introduced this approach, utilizing natural language sentences to describe historical observations and prediction targets, thereby transforming the forecasting task into a language generation format.
Building upon this framework, \cite{xue2022leveraging} leverages language foundation models such as BERT~\cite{bert} as the backbone network of the language generation part to conduct human mobility tasks.
The recent work~\cite{xue2023promptcast} further introduced the concept of PromptCast that extended language-based forecasting to other time series data domains like energy and temperature forecasting. However, the prompts used to describe time series are still based on fixed heuristic templates (\eg, similar to Figure~\ref{fig:intro} (b)). 

In this paper, we hypothesize that the performance of language-based forecasting is related to the ways of using language sentences to describe numerical data.
Unlike prior studies that have focused narrowly on fixed templates for prompting, our work aims to address this gap by exploring methods to develop more effective prompts, thereby potentially enhancing the performance of the language-based forecasting paradigm.

\begin{figure*}
    \centering
    \includegraphics[width=0.9\textwidth]{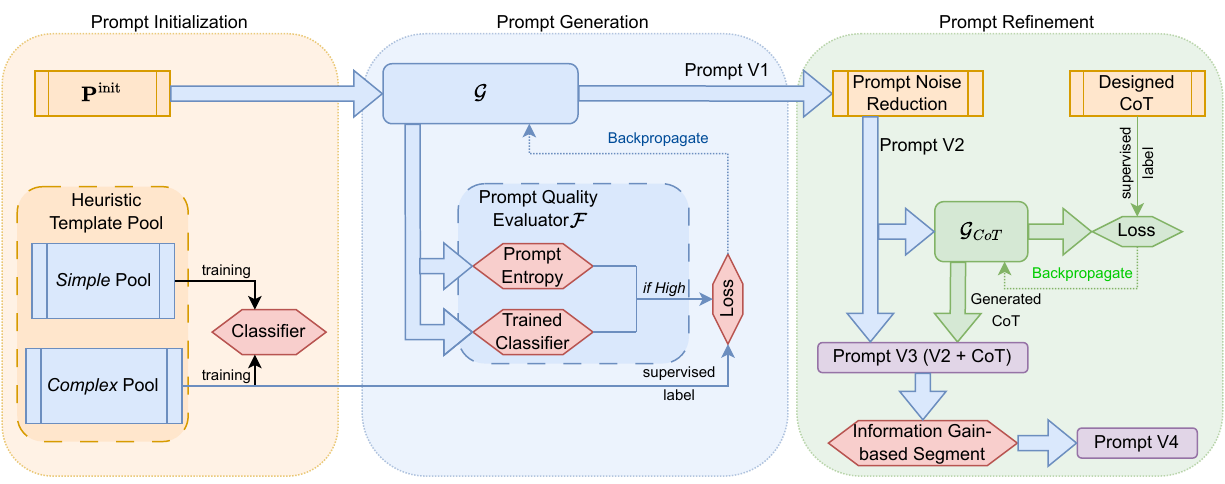}
    \caption{The illustration of our proposed prompt mining framework. It consists of 3 stages: prompt initialization, prompt generation, and prompt refinement.}
    \label{fig:framework}
\end{figure*}

\section{Method}

\subsection{Problem Formulation}
In this subsection, we formally define the problem that our proposed prompt mining framework aims to address: enhancing language-based human mobility forecasting through optimized prompt design.
As influenced by previous studies~\cite{xue2022translating,xue2022leveraging}, we simplify the forecasting task of interest as follows.
Given the historical customer visit records of a Point of Interest (POI) $m$ over a span of $n$ consecutive days, represented as $x^m_{t_1: t_{n}}=[x^m_{t_1}, x^m_{t_2}, \cdots, x^m_{t_n}]$, the objective is to predict the number of visits $x^m_{t_{n+1}}$ for the subsequent day $t_{n+1}$.
Normally, different types of auxiliary information of $\text{POI}_m$ such as the area region information $a^m$, the semantic information $c^m$, the opening information (\eg, business opening time $o^m$ and ending time $e^m$), or fine-grained hourly visits are often available, which can be used to assist the forecasting.
For the sake of clarity and simplification, we will omit the superscript $m$ (indicative of the POI index) from this point onward.

Template and Prompt: To leverage language models for the forecasting purpose, existing methods employ a predefined prompting template $\textbf{P}$ to convert the numerical data (\eg, $x^m_{t_1: t_{n}}$ and $x_{t_{n+1}}$) into sentences (these sentences are referred as prompts) that can be directly processed by language models.
The template consists of placeholders for the actual values from the data, resulting in sentences that present the mobility information in a human-readable format.
Specifically, the template consists of two essential parts:
the history prompt template $\textbf{P}_{\text{history}}$ that describes the historical values as the input sentences of the language models and the future prompt template $\textbf{P}_{\text{future}}$ that defines the sentences for the description of future observations.

In this study, we focus on the design of $\textbf{P}$ 
(especially the input parts $\textbf{P}_{\text{history}}$ of language models) 
and aim to explore diverse prompts that are contextually relevant, coherent, and effective in guiding language models to produce accurate forecasts. Through the investigation of prompts, we seek to enhance the overall forecasting accuracy and reliability of the language-based human mobility forecasting approach.
\subsection{Framework Overview}
As illustrated in Figure~\ref{fig:framework}, the proposed prompt mining framework consists of three key stages: prompt initialization, prompt generation, and prompt refinement.
\begin{itemize}
    \item Prompt Initialization: The mining process starts with prompt initialization. At this stage, we set the groundwork by creating a diverse pool of potential prompt templates manually. These templates can be used as supervision signals to mentor the following generation stage. 
    \item Prompt Generation: In the prompt generation phase, we employ a language model $\mathcal{G}$ as the core engine to learn how to generate ``better'' prompts. An evaluator module $\mathcal{F}$ is specifically designed to indicate the quality of generated prompts in the training process of $\mathcal{G}$.
    \item Prompt Refinement: The final stage of our framework is prompt refinement. We introduce several mechanisms to further improve the quality of generated prompts from the previous generation step.
\end{itemize}
By combining these three stages, our prompt mining framework ensures the systematic evolution of prompts from their initial conception to refined and optimized forms.
The following subsections delve into each stage of the framework, providing comprehensive insights into the proposed methodology for providing better and more dynamic prompts for the language-based forecasting process.

\begin{table*}[]
\centering
\caption{The example of our initial prompt template using in the prompt initialization stage.}
\label{tab:init}
\small
\begin{tabular}{l|l|p{0.8\textwidth}} \hline
\multicolumn{2}{l|}{\begin{tabular}[c]{@{}l@{}}Initial Prompt \\ ($\textbf{P}_{\text{history}}$)\end{tabular}} & \begin{tabular}[c]{@{}l@{}}In Region WI, Osseo, what is the daily human mobility of Mobil Store from Mon to Wed? \\ {[}0,0,0,0,0,0,0,0,1,0,1,1,2,0,2,1,0,0,0,0,1,0,0,0,0,0,0,0,1,0,0,1,1,0,0,1,2,1,1,0,0,0,1,1,0,1,0,1,0,0,0,0,0,1,0,0,1,1,1,1,0,0,1,1,1,0,3,0,0,1,0,0{].}\end{tabular} \\ \hline 
\end{tabular}
\end{table*}

\subsection{Prompt Initialization}
In our prompt mining framework, the initialization of prompts is a critical step to kickstart the process of generating high-quality forecasting prompts. The initial prompts serve as the foundation upon which subsequent iterations build, which also ensures the generated natural language has a certain directionality.

In a nutshell, the overarching objective of prompt mining is to use a designated language model $\mathcal{G}$ to automatically generate prompts from the original raw data, which is typically presented in numerical format.
However, as the first attempt, the absence of well-established datasets containing diverse prompts for human mobility data, along with a lack of relevant pre-trained language models for mobility prompt generation, poses a unique challenge.
Given these considerations, we introduce a pragmatic approach. Instead of directly generating prompts from the raw numerical mobility data, we propose to leverage the fixed template (referred to as the initial template) used in previous work as the input of the prompt generation model $\mathcal{G}$ and $\mathcal{G}$ is trained to yield better prompts from the initially given input throughout the prompt mining process.

\subsubsection{Initial Prompt Template}
As exampled in Table~\ref{tab:init}, we present a straightforward and uncomplicated template that functions as the initial template within our mining framework.
As aforementioned, this initial template serves as the foundation for transforming raw mobility data into the preliminary prompts denoted as $\textbf{P}^{\text{init}}$, which are considered as the starting prompts of the entire prompt mining process.
Thus, given an input instance $X, X \in \textbf{P}^{\text{init}}$, the prompt generation process (elaborated upon in Section~\ref{sec:p_gen}) can be succinctly expressed as:
\begin{equation}
    \hat{Y} = \mathcal{G}(X), \label{eq:1}
\end{equation}
where $\hat{Y}$ represents the generated prompts in the natural language format for the specified input instance.

\subsubsection{Templates Pool}\label{sec:pool}
Based on the initial template, the inputs of model $\mathcal{G}$ (\ie, $\textbf{P}^{\text{init}}$) are obtained. 
To effectively train the model $\mathcal{G}$, a collection of pseudo labels is essential to supervise the training process. 
To this end, we establish a pool of templates, as detailed in Appendix~\ref{sec:app}, Table~\ref{tab:pool}. These templates are crafted based on simple heuristics; for instance, \textit{Complex} templates encompass more detailed information such as specific Points of Interest (POI) details and POI working hours, whereas lower quality templates are more simplistic. The pool consists of 12 lower quality templates and 6 \textit{Complex} templates.

These templates support two main purposes: firstly, they are employed to train a classifier (refer to Section~\ref{sec:pqi}) capable of screening the quality (whether high or low) of a given prompt in the later prompt generation stage; secondly, they serve as pseudo labels during the training of $\mathcal{G}$.
It is worth noting that despite the discrepancy in the number of \textit{Simple} and \textit{Complex} templates, the generated training \textit{Simple} and \textit{Complex} prompts instances are equivalent in number and balanced.

Importantly, during the training process of $\mathcal{G}$, it is noteworthy that these heuristic templates are utilized only as pseudo labels and not as ground truth labels. 
This ensures that the model $\mathcal{G}$ generates dynamic prompts rather than replicating the exact prompts given in the template pool. The objective is to enable $\mathcal{G}$ to mine and discover prompts that are contextually sophisticated and better suited for human mobility forecasting.

\begin{table*}[]
\centering
\caption{The examples of our 4 prompt variants. The auxiliary context, temporal information, mobility data, and chain of thought parts are shown in orange, blue, red, and green.}
\label{tab:prompt}
\begin{tabular}{l|l|p{0.8\textwidth}} \hline
\multirow{2}{*}{V1} & $\textbf{P}_{\text{history}}$ & This is a {\color[HTML]{FFA500} Mobil in WI, Osseo}. The human mobility of the {\color[HTML]{0000FF}past 3 days} are: {\color[HTML]{FF0000} 0, 0, 0, 0, 0, 0, 0, 0, 1, 0, 1, 1, 2, 0, 2, 1, 0, 0, 0, 0, 1, 0, 0, 0} people (per hour) came here {\color[HTML]{FFA500}from 00:00 to 24:00 (working time)} on {\color[HTML]{0000FF}Mon}. {\color[HTML]{FF0000}0, 0, 0, 0, 1, 0, 0, 1, 1, 0, 0, 1, 2, 1, 1, 0, 0, 0, 1, 1, 0, 1, 0, 1} people (per hour) came here from {\color[HTML]{FFA500}from 00:00 to 24:00 (working time)} on {\color[HTML]{0000FF}Tue}. {\color[HTML]{FF0000}0, 0, 0, 0, 0, 1, 0, 0, 1, 1, 1, 1, 0, 0, 1, 1, 1, 0, 3, 0, 0, 1, 0, 0} people (per hour) came here {\color[HTML]{FFA500}from 00:00 to 24:00 (working time)} on {\color[HTML]{0000FF}Wed}. How many people will visit this place tomorrow? \\\cline{2-3}
 & $\textbf{P}_{\text{future}}$ & On {\color[HTML]{0000FF}Thu}, there are {\color[HTML]{FF0000}0, 1, 0, 1, 0, 0, 1, 2, 1, 0, 0, 1, 3, 0, 1, 2, 0, 1, 0, 1, 0, 0, 2, 0} people who will visit {\color[HTML]{FFA500}Mobil} during working time. \\ \hline \hline
\multirow{2}{*}{V2} & $\textbf{P}_{\text{history}}$ & This is a {\color[HTML]{FFA500}Mobil in WI, Osseo}. The human mobility of the {\color[HTML]{0000FF}past 3 days} are: {\color[HTML]{FF0000}5} people came here during the {\color[HTML]{FFA500}first half of the work shift} and {\color[HTML]{FF0000}4} people came here during the {\color[HTML]{FFA500}latter half of the work shift} on {\color[HTML]{0000FF}Mon}. {\color[HTML]{FF0000}4} people came here during the {\color[HTML]{FFA500}first half of the work shift} and {\color[HTML]{FF0000}8} people came here during the {\color[HTML]{FFA500}latter half of the work shift} on {\color[HTML]{0000FF}Tue}. {\color[HTML]{FF0000}5} people came here during the {\color[HTML]{FFA500}first half of the work shift} and {\color[HTML]{FF0000}7} people came here during the {\color[HTML]{FFA500}latter half of the work shift} on {\color[HTML]{0000FF}Wed}. How many people will visit this place tomorrow? \\\cline{2-3}
 & $\textbf{P}_{\text{future}}$ & On {\color[HTML]{0000FF}Thu}, there will be {\color[HTML]{FF0000}7} people to visit Mobil during the {\color[HTML]{FFA500}first half of the work shift} and {\color[HTML]{FF0000}10} people to visit Mobil during the {\color[HTML]{FFA500}latter half of the work shift}. Therefore, there are {\color[HTML]{FF0000}17} people will visit here. \\ \hline\hline
\multirow{2}{*}{V3} & $\textbf{P}_{\text{history}}$ & This is a {\color[HTML]{FFA500}Mobil in WI, Osseo}. From {\color[HTML]{0000FF}Mon to Wed}, the human mobility during the {\color[HTML]{FFA500}first and second half working time} are {\color[HTML]{FF0000}5, 4, 4, 8, 5, 7}. {\color[HTML]{008000}The entire working time is composed of the first half and the second half. Therefore, from Mon to Wed, the total human mobility are 5 + 4 = 9, 4 + 8 = 12, 5 + 7 = 12.} How many people will visit this place tomorrow? \\\cline{2-3}
 & $\textbf{P}_{\text{future}}$ & On {\color[HTML]{0000FF}Thu}, there will be {\color[HTML]{FF0000}7} people to visit {\color[HTML]{FFA500}Mobil} during the {\color[HTML]{FFA500}first half of the work shift} and {\color[HTML]{FF0000}10} people to visit {\color[HTML]{FFA500}Mobil} during the {\color[HTML]{FFA500}latter half of the work shift}. Therefore, there are {\color[HTML]{FF0000}17} people will visit here. \\\hline\hline
\multirow{2}{*}{V4} & $\textbf{P}_{\text{history}}$ & This is a {\color[HTML]{FFA500}Mobil in WI, Osseo}. From {\color[HTML]{0000FF}Mon to Wed}, the human mobility during the {\color[HTML]{FFA500}4 different time segments} are {\color[HTML]{FF0000}1, 4, 3, 1; 0, 5, 1, 6; 0, 6, 1, 5}. {\color[HTML]{008000}The entire working time is composed of the whole time segments. Therefore, from Mon to Wed, the total human mobility are 1 + 4 + 3 + 1 = 9; 0 + 5 + 1 + 6 = 12; 0 + 6 + 1 + 5 = 12.} How many people will visit this place tomorrow? \\\cline{2-3}
 & $\textbf{P}_{\text{future}}$ & On {\color[HTML]{0000FF}Thu}, there will be {\color[HTML]{FF0000}3, 5, 2, 7} people to visit {\color[HTML]{FFA500}Mobil} during these {\color[HTML]{FFA500}4 different time segments}. Therefore, there are {\color[HTML]{FF0000}17} people will visit {\color[HTML]{FFA500}Mobil} on {\color[HTML]{0000FF}Thu}.\\\hline
\end{tabular}
\end{table*}

\subsection{Prompt Generation}\label{sec:p_gen}
As illustrated in the middle module of Figure~\ref{fig:framework}, the prompt generation stage is the core part of the entire prompt mining process.
It functions based on the initial prompt as well as the heuristic template pool from the previous prompt initialization stage. 
The prompt generation stage consists of two key components: the Prompt Quality Evaluator $\mathcal{F}$ and the generation model $\mathcal{G}$.

\subsubsection{Prompt Quality Evaluator}\label{sec:pqi}
During the training process of model $\mathcal{G}$, it is important to inform the model of the quality of the generated prompts.
This motivates the introduction of a specialized module – the Prompt Quality Evaluator. 
This component plays a pivotal role in the prompt generation process, which serves as a critical feedback mechanism to assess the quality and effectiveness of generated prompts. It aids in distinguishing between high-quality and low-quality prompts, thereby guiding the iterative optimization of $\mathcal{G}$.

Based on the \textit{Simple}/\textit{Complex} templates in the template pool, we use a modest portion of (20\%) available numerical data to generate prompts. Specifically, we randomly select a template from the pool to make prompts and we make sure the total number of generated \textit{Complex} prompts is the same as \textit{Simple} prompts.
The central role of this classifier is to perform a binary classification, determining whether a given prompt $\hat{Y}$ aligns with a high-quality ($f(\hat{Y})=1$) or a low-quality ($f(\hat{Y})=0$) from the heuristic perspective.


Given the heuristics used to generate templates pool lack a concrete quantitative measure to gauge quality performance, a supplementary criterion is essential to fully discern the quality of generated prompts.
Essentially, the prompt generating is a process of information transmission, a translation from the domain of raw data into a descriptive language space.
Inspired by the observation that the language with higher information entropy is more efficient in transmitting information under different language systems~\cite{montemurro2011universal,liu2022entropy}, we introduce a special element to the Prompt Quality Evaluator – the Prompt Entropy.
Within the prompt context, a higher entropy signifies a greater level of uncertainty in the information encapsulated within the prompt. This enhanced uncertainty translates to a higher potential for the training of more effective forecasting models.

The information entropy of a yield prompt $H(\hat{Y})$ is calculated at the character level:
\begin{equation}
    H(\hat{Y}) = -\sum p(\gamma) \cdot log_2(p(\gamma))
\end{equation}
Here, $p(\gamma)$ represents the probability of the symbol or character $\gamma$ occurring within the prompt.
Thus, by synergistically integrating the rules established by the heuristic classifier and the prompt entropy metric, the ultimate prompt quality $\mathcal{F}(\hat{Y})$ is formulated as:
\begin{equation}
    \mathcal{F}(\hat{Y}) = f(\hat{Y}) \cdot \mathbbm{1}(H(\hat{Y}) \geqslant \tau), \label{eq:2}
\end{equation}
where $\mathbbm{1}$ is the indicator function and $\tau$ is the employed entropy threshold.

\subsubsection{Training of $\mathcal{G}$}
As expressed in Equation~\eqref{eq:1}, during the training process of $\mathcal{G}$, the input prompts stem from the initial template's generated prompts.
In this backpropagation process, the quality evaluator $\mathcal{F}$ plays a pivotal role in determining which output instances contribute to the training loss computation for the parameter update in $\mathcal{G}$.
The decision is guided by the conditions outlined in Equation~\eqref{eq:2}, where exclusively high-quality prompts are retained, while low-quality prompts are discarded. This selection mechanism enforces the focus on generating high-quality prompts, driving the iterative process towards improved language generation performance.

To enhance $\mathcal{G}$'s capacity to generate high-quality prompts, the \textit{Complex} pool established during prompt initialization is utilized as pseudo labels. Conceptually, once tokenized, both the pseudo label prompt $Y$ and the generated prompt $\hat{Y}$ can be represented as lists of tokens: $Y = [y_1, y_2, ..., y_j, ..., y_J]$ and $\hat{Y} = [\hat{y_1}, \hat{y_2}, ..., \hat{y_j}, ..., \hat{y_J}]$.
Furthermore, a novel loss function is introduced to actively guide $\mathcal{G}$ in generating high-quality prompts. The reciprocal of the associated prompt entropy is employed as the weight for the standard cross-entropy loss during training. In essence, this means that prompts with higher information entropy receive lower weights, resulting in smaller losses. This strategic approach accelerates model convergence and encourages a preference towards high-quality prompts during the training.
Consequently, the loss function underpinning our prompt generation process is formulated as:
\begin{equation}
    \mathcal{L}(\hat{Y}, Y) = \frac{\text{CrossEntropy}(\hat{Y}, Y)}{H(\hat{Y})} = - \frac{\sum_{j = 1}^J y_j log(\hat{y_i})}{H(\hat{Y})} \label{eq:loss}
\end{equation}
This concludes the generation stage. After training, the model $\mathcal{G}$ can be used to generate prompts from the initial template and the generated prompt is exemplified in Table~\ref{tab:prompt} (the V1 part).

\subsection{Prompt Refinement}
Building upon the above generation stage, wherein $\mathcal{G}$ is trained to generate prompts, the immediate prompts resulting from this stage are referred to as V1 prompts. In this section, our attention shifts to the refinement of the V1 prompts, with the aim of uncovering additional prompt variations.

\subsubsection{Prompt Noise Reduction}
The V1 prompts are designed to furnish an expanded array of information and finer-grained mobility data (\eg, including details such as hourly visit counts). However, the adoption of this strategy inevitably introduces a certain degree of noise into the generated prompts. This noise might manifest as extra characters or spacers placed between numbers.

Based on the V1 prompts, we introduce a subsequent iteration denoted as V2 prompts. The essence of V2 lies in its adoption of an information integration strategy that seeks to mitigate noise while retaining the core semantic content inherent in the V1 prompts.
Specifically, considering the overarching forecasting target as the aggregate number of visits for the subsequent day $t_{n+1}$ (same as the problem formulation in previous work~\cite{xue2022translating,xue2022leveraging}, V2 prompts reorganize human mobility data based on two distinct time periods. These divisions delineate the first and second halves of the day, effectively partitioning working hours at 12 PM, \ie, the diurnal partitioning.
This operational adjustment yields an updated set of prompts represented in the dedicated V2 section of Table~\ref{tab:prompt}.
Importantly, this refinement updates the prompt $\textbf{P}_{\text{history}}$, which further leads to a corresponding modification of $\textbf{P}_{\text{future}}$ (the prompt of the forecasting target).

\subsubsection{Chain of Thought Integration}
Inspired by the recent strides made in harnessing chain of thought applications within language models for enhancing reasoning capabilities~\cite{wei2022chain}, we extend our framework's refinement phase to incorporate a Chain of Thought (CoT) integration.
To this end, we propose the integration of a model $\mathcal{G}_{CoT}$ dedicated to generating a chain of thought, using the V2 prompts as input. In training or fine-tuning $\mathcal{G}_{CoT}$, we establish an initial format of the chain of thought as labels, effectively instructing $\mathcal{G}_{CoT}$ in its generation task.
Our devised CoT structure is conceptually grounded in the V2 prompts. This structure imparts the model with the understanding that the sum of mobility data during the first and second halves of working hours equates to the total daily human mobility count.

It is important to note that, distinct from $\mathcal{G}$'s training, $\mathcal{G}_{CoT}$ is trained using the standard cross-entropy loss, without employing prompt entropy. As depicted in Figure~\ref{fig:framework} (the green parts), the trained $\mathcal{G}_{CoT}$ generates a CoT, which is subsequently appended to the V2 prompts, leading to the formulation of the V3 prompts.
Table~\ref{tab:prompt} presents an illustrative example of V3 prompts.
From the example, we can also notice that the $\textbf{P}_{\text{future}}$ remains the same from V2 to V3 as the CoT is appended as the input of the language models when performing forecasting tasks. 
The sentences highlighted in green represent the CoT generated by $\mathcal{G}_{CoT}$. V3 prompts effectively merge human mobility data, eliminate redundancy, and establish a coherent chain of thought characterized by a fixed logical sequence of prompt sentences (\ie, the sentences in V2). 
This strategy augments the richness of valid information while concurrently attenuating potential noise. By introducing this CoT integration strategy, we enhance the sophistication of our prompt mining framework.

\subsubsection{Segmentation with Information Gain}
While the diurnal partitioning approach adopted in V2 and V3 effectively divides daily human mobility into two segments, this method may not fully capture the intricate patterns inherent in mobility behaviours. Such an approach might inadvertently result in the loss of crucial information, such as the peaks of the lunchtime rush around 12 PM.
In this phase of refinement, the focal challenge revolves around the effective segmentation of the given time series while minimizing the introduction of extraneous noise. The objective is to uncover a segmentation strategy that optimally captures temporal transitions in human activities and daily routines, ensuring accuracy while maintaining coherence.

To address this challenge, we draw upon an Information Gain-based Temporal Segmentation (IGTS) method~\cite{sadri2017information}), which is an unsupervised segmentation technique to find the transition times in human activities and daily routines.
In the pursuit of refined prompts (denoted as V4 prompts), IGTS is employed. 
Assuming that the mobility series (\eg, the numerical numbers) in one day is a list of $S$ and the goal is to divide it into $K$ segments $s_k, 1 \leqslant k \leqslant K$, with IGTS, this can be achieved by minimizing the information gain-based lost function $\mathcal{L}_{IGTS}(S)$:
\begin{equation}
    \mathcal{L}_{IGTS}(S)=H(S)-\sum_{k=1}^{K}\frac{\left | s_k \right |}{| S |}H(s_k),
\end{equation}
where $\left | s_k \right |$ is the length of the $k$-th segment (\ie, how many numbers) and $\left | S \right |$ is the total length of data to be segmented.

This approach contrasts with the binary halves used in diurnal partitioning within V2 and V3. As showcased in Table~\ref{tab:prompt}, the updated V4 prompts support multiple segments.
Considering the revisions to the segmentation in V4, both the CoT within $\textbf{P}_{\text{history}}$ and the $\textbf{P}_{\text{future}}$ are also modified accordingly.

\subsection{Summary of Designed Prompt Variants}
To provide a comprehensive comparative perspective on the four distinct prompt variants discovered through our prompt mining methodology, as illustrated in Table~\ref{tab:prompt}, we present a summary of the key properties of 4 prompt variants:
(1) V1: Generated directly by $\mathcal{G}$ which is trained with the prompt quality evaluator.
(2) V2: Explores diurnal partitioning for segmentation based on V1, focusing on the first and second halves of the day, which aims to reduce noise while preserving semantic content.
(3) V3: Integrates chain of thought (CoT) in $\textbf{P}_{\text{history}}$ based on V2.
(4) V4: Leverages Information Gain-based Temporal Segmentation (IGTS) to support detailed temporal segmentation beyond the diurnal partitioning in V2 and V3.


\section{Experiments}

\subsection{Source Data}
To validate our prompt mining approach for language-based mobility forecasting, we follow the previous language-based human mobility forecasting studies~\cite{xue2022translating,xue2022leveraging} and select SafeGraph\footnote{https://www.safegraph.com/} mobility pattern data accessed through Dewey research data platform\footnote{https://www.deweydata.io/} for our experiments.
We collect a total of 562,151 rows of raw weekly data of US POIs from 172 brand types.
The temporal span of the data ranges from 26th December 2022 to 2nd January 2023.
Each row provides one week of visiting data (including both daily counting and hourly counting) of one POI.
Auxiliary information such as the region of POI and the opening/closing hours are also available in the raw data.

\begin{table}[]
\centering
\caption{The results of numerical methods, using basic prompt and V1 generated by our prompt mining pipeline.}
\label{tab:res}
\small
\addtolength{\tabcolsep}{-0.75ex}
\begin{tabular}{l|l|cc|cc|cc} \hline
 &  & \multicolumn{2}{c|}{1st Half} & \multicolumn{2}{c|}{2nd Half} & \multicolumn{2}{c}{Daily Forecast} \\ \cline{3-8}
 &  & RMSE & MAE & RMSE & MAE & RMSE & MAE \\ \hline
 & ARIMA & 11.05 & 2.99 & 12.38 & 2.71 & 25.58 & 4.18 \\
 & Prophet & 92.69 & 35.49 & 9.25 & 3.02 & {\color[HTML]{0000FF} 10.76} & 3.89 \\
 & Autoformer & 25.53 & 4.9 & 10.6 & 3.79 & 35.02 & 8.31 \\
 & Crossformer & 24.77 & 3.69 & 9.81 & 3.05 & 33.68 & 6.42 \\
 & DLinear & 24.77 & 3.87 & 10.02 & 3.29 & 33.91 & 6.8 \\
 & Informer & 25.17 & 3.87 & 10.09 & 3.33 & 34.4 & 6.85 \\
 & LightTS & 24.47 & 4.03 & 9.97 & 3.44 & 33.51 & 7.04 \\
 & PatchTST & 24.36 & 3.69 & 9.98 & 3.08 & 33.42 & 6.39 \\
 & Pyraformer & 24.88 & 3.69 & 9.91 & 2.11 & 33.91 & 6.44 \\
 & Reformer & 25.08 & 3.85 & 10.02 & 3.2 & 34.24 & 6.7 \\
\multirow{-12}{*}{\begin{turn}{90}Numerical Forecasting\end{turn}} & TimesNet & 24.94 & 3.61 & 10.15 & 3.15 & 34.21 & 6.4 \\ \hline \hline
 & Bart & 57.46 & 15.7 & 58.78 & 56.43 & 94.54 & 65.05 \\
 & Bigbird & 13.3 & 4.61 & 10.6 & 4.15 & 22.9 & 8.71 \\ 
\multirow{-3}{*}{\begin{turn}{90}\begin{tabular}[c]{@{}c@{}}Basic \\ Prompt\end{tabular}\end{turn}} & Pegasus & 12.91 & 4.57 & 9.97 & 4.1 & 21.33 & 8.59 \\ \hline \hline
 & Bart & 11.01 & 3.38 & 22.92 & 20.51 & 23.43 & 19.71 \\
 & Bigbird & {\color[HTML]{0000FF}7.84} & {\color[HTML]{0000FF} 2.53} & {\color[HTML]{0000FF} 3.83} & {\color[HTML]{0000FF} 1.34} & {\color[HTML]{0000FF} 10.76} & {\color[HTML]{0000FF} 3.83} \\ 
\multirow{-3}{*}{\begin{turn}{90}\begin{tabular}[c]{@{}c@{}}Ours \\ V1\end{tabular}\end{turn}} & Pegasus & {\color[HTML]{FF0000} 5.77} & {\color[HTML]{FF0000} 2.42} & {\color[HTML]{FF0000} 3.58} & {\color[HTML]{FF0000} 1.31} & {\color[HTML]{FF0000} 8.26} & {\color[HTML]{FF0000} 3.67} \\ \hline
\end{tabular}
\end{table}

\subsection{Implementation Details}
In our experiments, the implementation includes two phases: the prompt mining phase and the forecasting phase.
For the prompt mining phase, 
we opted for the GPT-2 model~\cite{gpt2} architecture for both $\mathcal{G}$ and $\mathcal{G}_{CoT}$, despite the emergence of more advanced models like GPT-4~\cite{openai2023gpt4} and Llama 2~\cite{touvron2023llama}. The choice was driven by accessibility and computational feasibility considerations, as the latter models pose challenges in terms of availability and resource requirements.
The maximum length of the prompt generated by the GPT-2 model is limited to 512 tokens.
20\% of the data was allocated and passed to the templates given in the heuristic pool for the training of $\mathcal{G}$ and $\mathcal{G}_{CoT}$.
The batch size and dropout rate are set as 5 and 0.1, respectively. A learning rate initially is set at $5 \times 10^{-5}$ and early-stopping is adopted with a patience 3 epoch to avoid overfitting.

For the prompt quality evaluator, entropy analysis revealed that most generated prompts exhibited entropy values between 2 and 5. Setting a threshold $\tau$ at 3.5 enabled effective classification of high-quality and low-quality prompts based on their information entropy.
As for the classifier in the evaluator, given that the task is binary classification, a simple yet effective logistic regression classifier is selected. 
It uses L2 regularization and sets the reciprocal of the regularization strength as 1, which means that the strength is inversely proportional to the regularization. The model is fitted with the quasi-Newton as the optimization solver and a maximum iteration limit of 100.

In the forecasting phase, the total data is split into a training set (70\%, for training numerical forecasting methods and fine-tuning language models), a validation set (10\%), and a testing set (20\%).
The raw data is transformed into sentences by the different variants of developed prompts to fine-tune language models for forecasting purposes.
The fine-tuning process of the language forecasting models utilizes the standard Trainer provided by HuggingFace.
No changes are made to the loss function or other aspects during the fine-tuning of language models in this phase.
For the forecasting setting, we observe the mobility data of the last 3 days and predict the mobility of the next day. From the hourly data perspective, the input length is 72 hours and the forecasting horizon is 24 hours.
All of the experiments are performed with PyTorch on a Linux server equipped with Nvidia V100 GPUs and only one GPU was utilized.

\subsection{Evaluation}
The overall evaluation of different prompts is conducted based on the forecasting performance.
For language-based forecasting, following previous work~\cite{xue2022translating,xue2022leveraging}, we first need to extract the predicted numerical values from generated output sentences through string parsing.
After extracting the predicted numerical values from the generated sentences, the evaluation of the proposed method can be carried out similarly to how traditional numerical-based forecasting methods are evaluated.
We consider two error measures: the Root Mean Squared Error (RMSE) and the Mean Absolute Error (MAE). 

\subsection{Baselines}
In this subsection, we present the methods used for comparison in our energy load forecasting experiments. We categorize the baselines into two categories: numerical forecasting methods and language models.
The numerical baselines include the classic ARIMA, deep learning-based (Prophet~\cite{taylor2018forecasting}, TimesNet~\cite{wu2022timesnet}, LightTS~\cite{zhang2022less}, DLinear~\cite{zeng2023transformers}), and Transformer~\cite{vaswani2017attention}-based models ( Reformer~\cite{kitaev2019reformer}, Informer~\cite{zhou2021informer}, Autoformer~\cite{xu2021autoformer}, Pyraformer~\cite{liu2021pyraformer}, Crossformer~\cite{zhang2022crossformer}, PatchTST~\cite{nie2022time}). Except for ARIMA and Prophet, all the rest of the numerical methods are based on the implementation provided by the Time Series Library (TSlib)\footnote{\url{https://github.com/thuml/Time-Series-Library}.}.
For the language models, we select three language models with pre-trained weights from HuggingFace: Bart~\cite{bart}, Bigbird~\cite{zaheer2020big}, and Pegasus~\cite{zhang2020pegasus}.
From a recent benchmark study~\cite{xue2023promptcast}, these 3 models have shown good performance in forecasting time series compared to other language models.

\subsection{Comparison against Baselines}
In this part of the experiment, we focus on investigating the performance of our mined V1 prompts alongside various baselines.
The results of these evaluations are presented in Table~\ref{tab:res}.
For the numerical forecasting baselines, the inputs are the raw hourly visiting data in numbers.
For the language models, we also compare V1 against a basic prompt template (\eg, This POI is [STORE TYPE]. There were [HOURLY VISITING DATA] people came here to visit.) drawn from the previous language-based forecasting study~\cite{xue2023promptcast}.
All methods reported in Table~\ref{tab:res} predict hourly forecasting during the entire working day.
In addition to the daily forecasting (the sum of hourly predictions) performance, to make a more accurate comparison with the subsequent generated prompt V2 and V3 (where the diurnal partitioning is applied) later, we also report RMSE and MAE within the first and second-half of the working time. It allows better measurement of their predictive performance over different time periods.
The best and the second-best performers under each column are shown in red and blue.

In general, our V1 outperforms both numerical forecasting methods and the basic prompt without prompt mining process.
While numerical models show competitive results in the second half, the Pegasus model with our V1 prompts consistently maintains a superior performance across different forecasting scenarios.
For the same language model, we can see that using V1 leads to better forecasting performance than using the basic prompt.
Moreover, for language models, regardless of the prompt employed, both Pegasus and Bigbird showcase superior forecasting performance compared to Bart. 
The above comparison results validate the
effectiveness of our proposed prompt mining method that can contribute to more accurate forecasting.

\subsection{Performance of Different Prompt Variants}
To have a deeper understanding of our prompt mining approach and its impact on forecasting accuracy, we conduct a detailed analysis of the performance of different prompt variants.
Table~\ref{tab:variant_res} presents the forecasting performance of the three language models using Prompt V1-V3. 

In general, we can observe that V3 outperforms V1 and V2 in terms of the daily forecasting results no matter which model is used.
For the Bart model, V2 exhibits substantial improvements in accuracy compared to V1, with significant reductions in both RMSE and MAE. V3 also has a large reduction in daily forecast RMSE.
In the Bigbird case, V2 and V3 display a remarkable improvement against V1 for the first half day. Although V1 has a better performance in the second half day, V3 still remains the top performer under the daily forecast.
Similarly, the outcomes observed for the Pegasus language model are almost the same as the trends apparent in the Bigbird analysis. 
This analysis of different prompt variants shows the efficacy of our mining method in enhancing mobility forecasting accuracy. The prompt refinements introduced in the prompt mining process, particularly for V3, contribute to improvements in the forecasting performance, which justifies the significance of our approach's prompt refinement phase.

\begin{table}[]
\centering
\caption{The results of different prompt variants.}
\label{tab:variant_res}
\addtolength{\tabcolsep}{-0.35ex}
\begin{tabular}{c|c|cc|cc|cc} \hline 
\multicolumn{1}{l}{} &  & \multicolumn{2}{c|}{1st Half} & \multicolumn{2}{c|}{2nd Half} & \multicolumn{2}{c}{Daily Forecast} \\ \cline{3-8}
\multicolumn{1}{l}{} & \multirow{-2}{*}{Prompt} & RMSE & MAE & RMSE & MAE & RMSE & MAE \\ \hline 
 & V1 & 11.01 & 3.38 & 22.92 & 20.51 & 23.43 & 19.71 \\
 & V2 & 3.75 & 1.08 & 4.71 & 1.61 & 8.79 & 2.82 \\
\multirow{-3}{*}{\begin{turn}{90}Bart\end{turn}} & V3 & 3.19 & 1.08 & 4.48 & 1.62 & {\color[HTML]{0000FF} 6.68} & 2.42 \\ \hline \hline 
 & V1 & 7.84 & 2.53 & {\color[HTML]{0000FF} 3.83} & {\color[HTML]{0000FF} 1.34} & 10.76 & 3.83 \\
 & V2 & {\color[HTML]{FF0000} 2.65} & 0.99 & 7.46 & 1.68 & 8.87 & 2.37 \\ 
\multirow{-3}{*}{\begin{turn}{90}Bigbird\end{turn}} & V3 & {\color[HTML]{0000FF} 2.78} & 1.00 & 4.47 & 1.59 & {\color[HTML]{FF0000} 6.00} & 2.29 \\ \hline \hline 
 & V1 & 5.77 & 2.42 & {\color[HTML]{FF0000} 3.58} & {\color[HTML]{FF0000} 1.31} & 8.26 & 3.67 \\
 & V2 & 2.97 & {\color[HTML]{FF0000} 0.97} & 5.19 & 1.54 & 8.58 & {\color[HTML]{0000FF} 2.21} \\
\multirow{-3}{*}{\begin{turn}{90}Pegasus\end{turn}} & V3 & 3.65 & {\color[HTML]{0000FF} 0.98} & 4.19 & 1.52 & 8.13 & {\color[HTML]{FF0000} 2.20} \\ \hline 
\end{tabular}
\end{table}

\subsection{Impact of Segments}

\begin{table}[]
\centering
\caption{Results of using different segments.}
\label{tab:seg_res}
\small
\addtolength{\tabcolsep}{-0.85ex}
\begin{tabular}{l|c|c|cc|cc} \hline 
Language &  &  & \multicolumn{2}{c|}{Segment Average} & \multicolumn{2}{c}{Daily Forecast} \\ \cline{4-7}
Model & \multirow{-2}{*}{Prompt} & \multirow{-2}{*}{$K$} & RMSE & MAE & RMSE & MAE \\ \hline 
 & V3 & default (2) & 3.84 & 1.35 & {\color[HTML]{0000FF} 6.68} & 2.42 \\ \cline{2-7}
 &  & 2 & 8.49 & 2.06 & 12.67 & 2.51 \\
Bart &  & 3 & {\color[HTML]{0000FF} 3.35} & 0.88 & 8.07 & {\color[HTML]{FF0000} 2.18} \\
 &  & 4 & {\color[HTML]{FF0000} 2.64} & {\color[HTML]{FF0000} 0.75} & 9.20 & {\color[HTML]{0000FF} 2.29} \\
 & \multirow{-4}{*}{V4} & 5 & 5.54 & {\color[HTML]{0000FF} 0.77} & {\color[HTML]{FF0000} 5.04} & 2.50 \\ \hline \hline 
 & V3 & default (2) & 3.63 & 1.30 & {\color[HTML]{0000FF} 6.00} & {\color[HTML]{0000FF} 2.29} \\ \cline{2-7}
 &  & 2 & 19.44 & 2.02 & 11.20 & 2.32 \\
Bigbird &  & 3 & 2.56 & 0.81 & {\color[HTML]{FF0000} 5.43} & {\color[HTML]{FF0000} 2.08} \\
 &  & 4 & {\color[HTML]{0000FF} 2.52} & {\color[HTML]{0000FF} 0.69} & 7.13 & {\color[HTML]{0000FF} 2.29} \\
 & \multirow{-4}{*}{V4} & 5 & {\color[HTML]{FF0000} 1.81} & {\color[HTML]{FF0000} 0.59} & 7.39 & 2.42 \\ \hline \hline 
 & V3 & default (2) & 3.92 & 1.25 & 8.13 & 2.20 \\ \cline{2-7}
 &  & 2 & 9.04 & 1.98 & 11.22 & 2.24 \\
Pegasus &  & 3 & 2.41 & 0.79 & {\color[HTML]{0000FF} 5.20} & {\color[HTML]{FF0000} 2.02} \\
 &  & 4 & {\color[HTML]{0000FF} 2.27} & {\color[HTML]{0000FF} 0.67} & {\color[HTML]{0000FF} 5.20} & {\color[HTML]{0000FF} 2.17} \\
 & \multirow{-4}{*}{V4} & 5 & {\color[HTML]{FF0000} 1.70} & {\color[HTML]{FF0000} 0.57} & {\color[HTML]{FF0000} 5.03} & 2.30 \\ \hline 
\end{tabular}
\end{table}

In this part of the experiment, we explore the impact of different segmenting mechanisms used in our prompt refinement stage.
Specifically, we investigate how different numbers of segments ($K$) in our V4 prompts influence forecasting performance. 
The results of this experimentation are summarized in Table~\ref{tab:seg_res}, where we also provide the forecasting performance of V3 for easy comparison, which employs diurnal partitioning with 2 segments.
Different from Table~\ref{tab:res} and Table~\ref{tab:variant_res}, we calculate the average RMSE/MAE of each segment instead of the first/second half RMSE/MAE since we have settings with more than 2 segments for V4.
We also highlight the best and second-best results for each language model separately.

Upon analyzing the results, it is evident that the performance of our V4 prompts is influenced by the number of segments $K$. 
For example, in the case of segment average performance, increasing the number of segments from 2 to 5 (V4) leads to a decreased error for both Bigbird and Pegasus. 
For the daily forecasting performance, a similar trend can also be noticed for the Bart and Pegasus models.
Moreover, the comparison with V3 prompts highlights that our refined V4 prompts generally perform favorably.
In conclusion, the results indicate that with a dynamic segmentation strategy offered in our V4 prompts, the language-based forecasting methods can be further improved. Additionally, we observe similar performance on different language models, which demonstrates the adaptability of our prompt mining approach.

\section{Conclusion}
In this study, we present a novel prompt mining framework for enhancing language-based mobility forecasting.
We introduce a multi-stage process with a core prompt generation stage based on prompt entropy and a refinement stage with the consideration of a chain of thought.
Our mining pipeline generates and refines prompts for transforming human mobility data into sentences in order to leverage language models for mobility forecasting.
Through comprehensive experiments, we demonstrate the superiority of our refined prompts over traditional numerical forecasting methods and basic prompt structures. Our results also highlight the consistent performance improvement across different language models used for the forecasting phase.
The future directions of mobility prompt mining can include the integration of external data sources, such as weather and events, which could further provide more context and enhance forecasting capabilities. How to involve human feedback from domain experts in the prompt mining process would be another interesting future research topic.



\begin{acks}
We acknowledge the support of Cisco Research Gift (CG\# 75677887), and the resources and services from the National Computational Infrastructure (NCI) that is supported by the Australian Government.
\end{acks}
\bibliographystyle{ACM-Reference-Format}
\bibliography{main}


\appendix

\begin{table*}[!ht]
\centering
\caption{The heuristic template pool includes 12 \textit{Simple} templates and 6 \textit{Complex} templates.}
\label{tab:pool}
\small 
\begin{tabular}{l|p{.9\textwidth}} \hline
                              & Template                                                \\ \hline
\multirow{12}{*}{\begin{turn}{90}\textit{Simple}\end{turn}} & A private store. Human mobility in past \{$n$\} days are: $x_{t_1: t_{n}}$. How many people will visit tomorrow?                                                                                                                                  \\\cline{2-2}
                              & A private store. There are $x_{t_1: t_{n}}$ came in the past \{$n$\} days, and how many people?                                \\\cline{2-2}
                              & A private store. There are $x_{t_1: t_{n}}$ came in the past \{$n$\} days. How many?                                           \\\cline{2-2}
                              & This is a private store in \{$a$\}. The human mobility of the past \{$n$\} days are: \{$x$\} people (per hour) visited in \{$o$\} - \{$e$\} on \{$t$\}-th day. {[}Repeat for $n$ days{]}. How many people will visit this place tomorrow?      \\\cline{2-2}
                              & How many people will visit this private store tomorrow? If the human mobility of the past \{$n$\} days are: \{$x$\} people (per hour) visited in \{$o$\} - \{$e$\} on \{$t$\}-th day. {[}Repeat for $n$ days{]}.                               \\\cline{2-2}
                              & In \{$a$\}, there is a private store and the human mobility of the past \{$n$\} days are: \{$x$\} people (per hour) visited in \{$o$\} - \{$e$\} on \{$t$\}-th day. {[}Repeat for $n$ days{]}. How many people will visit this place on tomorrow? \\\cline{2-2}
                              & \{$c$\}. Human mobility in past \{$n$\} days are: $x_{t_1: t_{n}}$. How many people will visit tomorrow?                                                                    \\\cline{2-2}
                              & \{$c$\}. There are $x_{t_1: t_{n}}$ came in the past \{$n$\} days, and how many people?                                     \\ \cline{2-2}
                              & \{$c$\}. There are $x_{t_1: t_{n}}$ came in the past \{$n$\} days. How many?                                                  \\ \cline{2-2}
                              & This is a \{$c$\} in \{$a$\}. The human mobility of the past \{$n$\} days are: \{$x$\} people (per hour) visited in \{$o$\} - \{$e$\} on \{$t$\}-th day. {[}Repeat for $n$ days{]}. How many people will visit this place tomorrow?            \\ \cline{2-2}
                              & How many people will visit \{$c$\} on tomorrow? If the human mobility of the past \{$n$\} days are: \{$x$\} people (per hour) visited in \{$o$\} - \{$e$\} on \{$t$\}-th day. {[}Repeat for $n$ days{]}.                                          \\ \cline{2-2}
                              & In \{$a$\}, there is a \{$c$\} and the human mobility of the past \{$n$\} days are: \{$x$\} people (per hour) visited in \{$o$\} - \{$e$\} on \{$t$\}-th day. {[}Repeat for $n$ days{]}. How many people will visit this place tomorrow?       \\ \hline \hline
\multirow{6}{*}{\begin{turn}{90}\textit{Complex}\end{turn}} & This is a private store in \{$a$\}. The human mobility of the past \{$n$\} days are: \{$x$\} people (per hour) visited in \{$o$\} - \{$e$\} on \{$t$\}-th day. {[}Repeat for $n$ days{]}. How many people will visit this place tomorrow?      \\ \cline{2-2}
                              & In \{$a$\}, how many people will visit this private store tomorrow? If the human mobility of the past \{$n$\} days are: \{$x$\} people (per hour) visited in \{$o$\} - \{$e$\} on \{$t$\}-th day. {[}Repeat for $n$ days{]}.                   \\ \cline{2-2}
                              & In \{$a$\}, there is a private store and the human mobility of the past \{$n$\} days are: \{$x$\} people (per hour) visited in \{$o$\} - \{$e$\} on \{$t$\}-th day. {[}Repeat for $n$ days{]}. How many people will visit this place on tomorrow? \\ \cline{2-2}
                              & This is a \{$c$\} in \{$a$\}. The human mobility of the past \{$n$\} days are: \{$x$\} people (per hour) visited in \{$o$\} - \{$e$\} on \{$t$\}-th day. {[}Repeat for $n$ days{]}. How many people will visit this place tomorrow?            \\ \cline{2-2}
                              & In \{$a$\}, how many people will visit \{$c$\} on tomorrow? If the human mobility of the past \{$n$\} days are: \{$x$\} people (per hour) visited in \{$o$\} - \{$e$\} on \{$t$\}-th day. {[}Repeat for $n$ days{]}.                              \\ \cline{2-2}
                              & In \{$a$\}, there is a \{$c$\} and the human mobility of the past \{$n$\} days are: \{$x$\} people (per hour) visited in \{$o$\} - \{$e$\} on \{$t$\}-th day. {[}Repeat for $n$ days{]}. How many people will visit this place tomorrow? \\ \hline     
\end{tabular}
\end{table*}

\section{Heuristic Prompt Template Pool}\label{sec:app}
In this section, we show the details of our heuristic template pool (described in Section~\ref{sec:pool}). As presented in Table~\ref{tab:pool}, this heuristic pool consists of both \textit{Simple} templates and \textit{Complex} templates.
\textit{Complex} templates, as opposed to their \textit{Simple} counterparts, offer more comprehensive information, featuring complete sentences to describe the semantic details of Points of Interest (POIs). These templates are instrumental in generating training prompt instances to facilitate the training of $\mathcal{G}$.

During the formation of the training set, these templates are randomly selected to create either \textit{Complex} or \textit{Simple} instances (prompts). To ensure a balanced dataset for training $\mathcal{G}$, the number of \textit{Complex}instances matches that of \textit{Simple} instances within the formed training set.




\end{document}